\documentclass[nocrop]{bioinfo}
\copyrightyear{2021} \pubyear{2021}

\usepackage{booktabs}
\usepackage{threeparttable}  
\usepackage{multirow}
\usepackage[normalem]{ulem}
\useunder{\uline}{\ul}{}
\usepackage{lscape} 
\usepackage{graphicx}
\usepackage[table,xcdraw]{xcolor}
\usepackage[normalem]{ulem}
\useunder{\uline}{\ul}{}
\usepackage{booktabs}
\usepackage{multirow}

\access{Advance Access Publication Date: Day Month Year}
\appnotes{Manuscript Category}

\begin{document}
\firstpage{1}
\subtitle{Data and text mining}

\title[SciFive: a text-to-text transformer model for biomedical literature]{SciFive: a text-to-text transformer model \\
for biomedical literature}

\author[Phan \textit{et~al}.]{Long N. Phan\,$^{1,2,^\dagger}$, James T. Anibal\,$^{2,^{\dagger}}$, Hieu Tran\,$^{3}$, Shaurya Chanana\,$^{4}$, \\ Erol Bahadıroğlu\,$^{2}$, Alec Peltekian\,$^{1,2}$, Grégoire Altan-Bonnet\,$^{2}$}
\address{$^{\text{\sf 1}}$Department of Computer Sciences, Case Western University, Cleveland OH, USA.\\
$^{\text{\sf 2}}$ImmunoDynamics Section, Laboratory of Integrative Cancer Immunology, National Cancer Institute, Bethesda MD, USA.\\
$^{\text{\sf 3}}$University of Science, Vietnam National University, Ho Chi Minh City, Vietnam.\\
$^{\text{\sf 4}}$Natural Products Branch, National Cancer Institute, Bethesda MD, USA.}

\corresp{$^\dagger$These authors contributed equally to this work.}

\abstract{\textbf{Motivation:} In 2019, researchers from Google released the Text-to-Text Transfer Transformer (T5) trained on the "Colossal Clean Crawled Corpus" (C4) This approach achieved state-of-the-art (SOTA) results on a diverse range of tasks related to natural language processing (NLP). In the last decade, NLP in biomedicine has become more prominent (i.e. text mining of scientific literature, analysis of electronic health records). This development has created a need for NLP methods trained on corpora of biomedical literature containing the dense technical language characteristic of scientific writing. In this report, we introduce a T5-based model that has been successfully shifted into the biomedical domain. \\
\textbf{Results:} In this report, we introduce SciFive, a domain-specific T5 model that has been pre-trained on large biomedical corpora. Our model outperforms the current SOTA methods (i.e. BERT, BioBERT, Base T5) on tasks in named entity relation, relation extraction, natural language inference, and question-answering. We show that text-generation methods have significant potential in a broad array of biomedical NLP tasks, particularly those requiring longer, more complex outputs. Our results support further research into biomedical text generation and the development of new methods in this area.\\
\textbf{Availability:} All checkpoints and pre-trained weights of SciFive are publicly available at https://console.cloud.google.com/storage/browser/scifive. The sources code for self-supervised and fine-tuned models is in https://github.com/justinphan3110/SciFive  \\
\textbf{Contact:} \href{ gregoire.altan-bonnet@nih.gov}{ gregoire.altan-bonnet@nih.gov}}

\history{Received on XXXXX; revised on XXXXX; accepted on XXXXX}

\editor{Associate Editor: XXXXXXX}

\maketitle

\section{Introduction}
Biomedical literature is widely accessible to the scientific community through databases such as Pubmed, PMC, and ScienceDirect. Within seconds, researchers can access millions of journal articles relating to an input query. Text generation tasks such as document summarization and question answering can allow researchers to quickly obtain important information from a large collection of papers, yet current methods generally underperform in these areas. Thus, new NLP methods are needed to parse the increasingly immense amounts of information.
\subsection{Related Work}
The introduction of the transformer \citep{DBLP:journals/corr/VaswaniSPUJGKP17} marked a significant achievement for natural language processing. This is demonstrated by the success of transformer-based architectures such as BERT \citep{DBLP:journals/corr/abs-1810-04805}, which, at the time of publication, achieved state-of-the-art (SOTA) results on common NLP tasks. Furthermore, the BERT model has been extended for domain-specific tasks in NLP. Domain-specific language (i.e. biomedical language) is often challenging for NLP models because of the significant differences in vocabulary compared to standard langauge corpora such as Wikipedia. To solve this problem, BERT models have been pre-trained for domain-specific tasks. With this approach, SOTA results were achieved in areas such as clinical notes, biomedical literature, and general scientific literature.

\section{Approach}
BERT \citep{DBLP:journals/corr/abs-1810-04805} is not a unified transfer learning method because BERT-style models can only produce a single prediction for a given input. These models are simply not designed for text generation tasks such as question-answering or summarization.  The text-to-text transfer transformer (T5) model proposed by \cite{DBLP:journals/corr/abs-1910-10683} overcomes this limitation by outputting a string of text for each input, allowing for both question-answering, summarization and other tasks where a single output is generally insufficient. In this report, we introduce SciFive, a pretrained, domain-specific adaptation of the T5 model that is intended for tasks relating to biomedical literature. We here outline two primary contributions of our work. 

(1) Our model achieves SOTA results on a variety of common classification tasks in biomedical NLP, including named entity recognition (NER) and relation-extraction (RE). 
 
 (2) Second, our model can be extended to tasks requiring extended outputs and achieves superior results on BioAsq question-answering challenges when compared to BioBERT, the current SOTA method to the best of our knowledge \citep{DBLP:journals/corr/abs-1901-08746}

\section{Unlabeled Dataset}
In this section we will describe our biomedical unlabeled datasets which are used in the transfer learning pre-training stage. These large datasets overcome the drawback of overfitting when building a language model in the biomedical domain.  \citep{DBLP:journals/corr/Ruder17a}. For SciFive, we use two different corpora of biomedical language in order to generalize our model within the domain.

\textbf{PubMed Abstract} \footnote{https://pubmed.ncbi.nlm.nih.gov}: The PubMed database contains more than 32 millions citations and abstracts of biomedical literature. For the purpose of model pre-training, we use only the abstracts.

\textbf{PubMed Central (PMC)} \footnote{https://www.ncbi.nlm.nih.gov/pmc}: PMC is a corpus of free full-text articles in the domain of biomedical and life sciences. We hypothesize that training the language model with full-text articles can improve the learning in biomedical context while still containing a generalized representation of natural language overall.

\section{Methods}
\begin{methods}
Here, we describe our approach to implementing the SciFive model, which retains the original structure and parameters of the T5 model \citep{DBLP:journals/corr/abs-1910-10683}.

\subsection{T5}
The text-to-text transfer transformer (T5) model \citep{DBLP:journals/corr/abs-1910-10683} is highly similar to the transformer-based encoder-decoder model introduced by \cite{DBLP:journals/corr/VaswaniSPUJGKP17}. Each encoder block consists of a self-attention layer and a feed-forward neural network. Each decoder block consists of a self-attention layer, an encoder-decoder attention layer, and a feedforward neural network. There are, however, minor differences between T5 and the transformer-based encoder-decoder model. For example, layer normalization is applied between the components of each encoder block and each decoder block. Compared to BERT \citep{DBLP:journals/corr/abs-1810-04805}, the addition of the decoder block allows T5 to generate outputs that are sequences of text. T5 is pre-trained with self-supervision through a learning objective called span-based language masking. \citep{DBLP:journals/corr/abs-1910-10683}.

\subsection{SciFive}
SciFive follows the sequence-to-sequence encoder-decoder architecture proposed by \cite{DBLP:journals/corr/VaswaniSPUJGKP17} and the T5 framework \footnote{https://github.com/google-research/text-to-text-transfer-transformer} released by \cite{DBLP:journals/corr/abs-1910-10683}. The original T5 work implemented five different model sizes - Small, Base, Large, 3B, and 11B. Due to limited computing resources, we will use only the base and large model for this study. The base and large models have 220 million parameters and 770 million parameters respectively.

\begin{table}[h]
\centering
\caption{Corpus combinations for SciFive}
\begin{tabular}{@{}ll@{}}
\toprule
Model                & Corpus Combination \\ \midrule
T5 \cite{DBLP:journals/corr/abs-1910-10683}                 & C4                 \\
SciFive(+pubmed)     & C4+pubmed          \\
SciFive(+pmc)        & C4+pmc             \\
SciFive(+pubmed+pmc) & C4+pubmed+pmc      \\ \bottomrule
\end{tabular}
\label{corpus_combination}
\end{table}
We first initialized SciFive with the pre-trained weights from the base T5 model. We then re-trained SciFive on various combinations of the C4 corpus \citep{DBLP:journals/corr/abs-2104-08758}, a corpus of PubMed abstracts, and a corpus of PMC full-text articles. We trained SciFive for extra 200k steps to optimize the pre-trained weights from T5 in the context of biomedical literature. We also trained a large version of the SciFive model, using 1.2 millions steps (200k additional steps compared to the regular model). With the provided TPU v2-8 on Google Colab, we used the self-supervised training setting recommended by \cite{DBLP:journals/corr/abs-1910-10683} with a batch size of 128 for the base model and 64 for the large model. We used a learning rate of 0.001 and sequence length 1024 tokens for both input and target as we noticed that unlabeled biomedical text during self-supervised training is long. For the purpose of generalization of biomedical text, we train SciFive on various combinations of biomedical corpus as describe in Table \ref{corpus_combination}.

\subsection{Input/Output Representation}
\begin{figure}[hbt!]
    \centering
    \includegraphics[width=0.45\textwidth,height=\textheight,keepaspectratio]{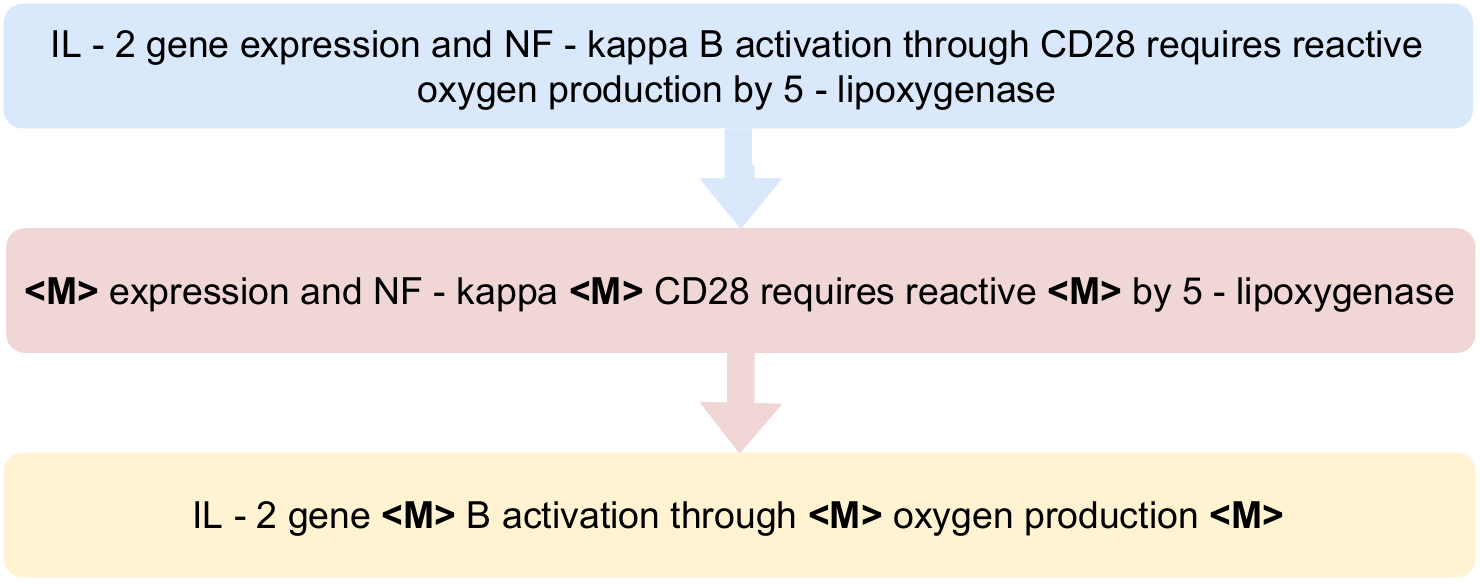}
    \caption{An illustration on Span-based mask language modeling. For the input sentence, the set of tokens "IL","-","2", "kappa", "B",..."oxygen", "production" is randomly chosen for corruption, where consecutive tokens are counted as spans and replaced by a sentinel unique masked token <M>. The output sequence then consists of the concatenation of the dropped-out spans, sentinel tokens used to replace them in the input and the final sentinel token. }
    \label{fig:learning_objective}
\end{figure}
Consistent with the original T5 model \cite{DBLP:journals/corr/abs-1910-10683}, SciFive converts all of the biomedical tasks into a text-to-text format. During self-supervised training, a text input sequence is given and the model will try to learn a target input going through a learning objective called span-based mask language modeling. Spans of text are randomly masked and the target sequence is predicted as a concatenation of the same sentinel tokens and the real masked spans. An illustration of span-based mask learning objective is in Figure \ref{fig:learning_objective}.

During supervised training, a sequence of text for both input and target is given to the model for the purpose of learning to generate text. For example, when performing Named-entity recognition (NER), we generate the target sequence by prepending and appending a special token to the named entities in a sentence. The target sequence for Question Answering task is the text corresponding to the answer for a given question (the question text is the input). 

\subsection{Vocabulary}
For every pre-trained language model (LMs), vocabulary plays a crucial role, as these models attempt to derive effective contextualized word vector representations from the training corpus. For SciFive, we use the Sentence Piece model \citep{DBLP:journals/corr/abs-1808-06226} as a base vocabulary model. Sentence Piece is used in all of our  SciFive models because it extracts sub-words that contain the semantic meaning of a sequence. This overcomes the drawbacks of word-level tokenization and eliminates the need for an immense vocabulary set. 

\subsection{Multi-Task Learning}
SciFive is trained with a maximum likelihood objective using "teacher forcing" \citep{DBLP:journals/corr/abs-1910-10683} for all tasks, thereby enabling multi-task learning. During supervised fine-tuning, a task-specific token is prepended to the input sequence. In one example, we leverage this type of training for the Named-entity recognition task. We believe that this strategy will boost performance for biomedical NER by using the attention of each named entity across all the tasks. Figure \ref{fig:multi_task_learning} illustrates multi-task learning for our NER tasks.

\begin{figure*}
    \centering
    \includegraphics[width=0.8\textwidth,height=18em]{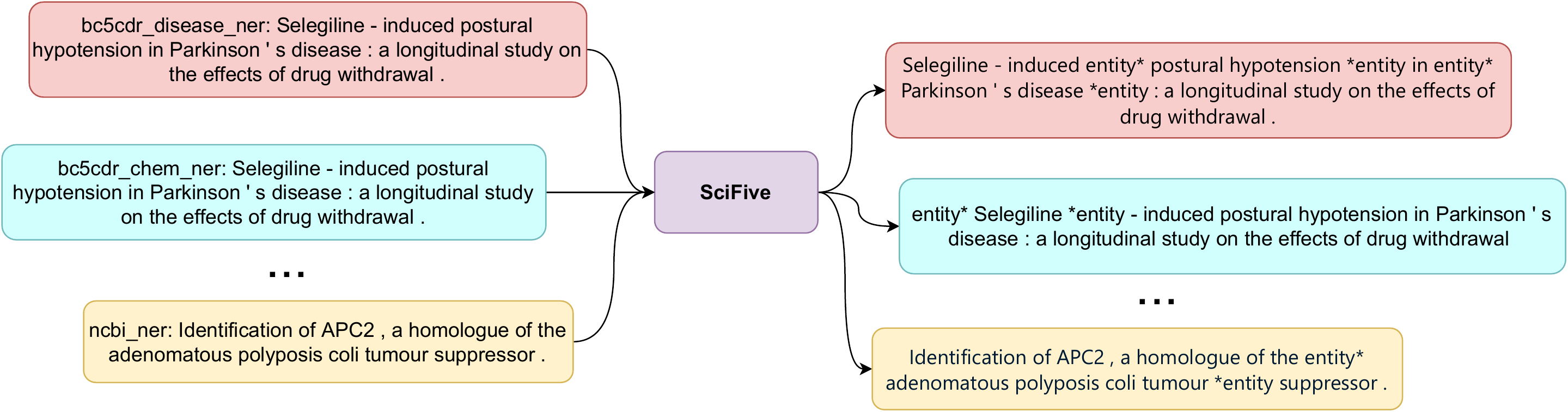}
    \caption{An illustration about Multi-task learning in Name-entity Recognition Tasks}
    \label{fig:multi_task_learning}
\end{figure*}

\subsection{Fine-Tuning SciFive}
We fine-tuned SciFive on five categories of biomedical NLP tasks.

(1)   Named entity recognition (NER)  involves predicting a predefined \indent category that describes a proper noun. For example “Lupus” may be \indent classified as “Disease".

(2)   Relation Extraction (RE) involves identifying relationships within \indent text           (i.e. gene-disease).

(3)   Natural Language Inference involves determining the         validity of a \indent hypothesis (i.e., True, False).

(4)   Document Classification involves assigning a document       to a \indent category based on the text.

(5)   Question answering involves generating an answer if         given a \indent question and a sequence of text containing the answer to that question. \\
 
We fine-tuned in both multi-tasking and single-task learning using the final checkpoints of our SciFive model, 200k steps for both base and large models. Similar to the setting during self-supervised training on TPU v2-8, we choose the batch size of 128 and 64 for the base and large respectively with learning rate 0.001. The input and output specification setting for each task is described in Table \ref{setting}.

\end{methods}
\begin{table*}[]
\centering
\caption{The input and target sequence length settings for each Self-supervised Learning, Name-entity Recognition, Relational Extraction, and Question Answering task}

\begin{tabular}{@{}clllccc@{}}
\toprule
Task                                     & Dataset         & Entity type          & Number of entities   & Task Type                             & Input Length         & Target Length        \\ \midrule
\multirow{3}{*}{Self-Supervise Learning} & PubMed          & \multicolumn{1}{r}{} & \multicolumn{1}{r}{} & \multicolumn{1}{r}{\multirow{3}{*}{}} & 1024                 & 1024                 \\
                                         & PMC             & \multicolumn{1}{r}{} & \multicolumn{1}{r}{} & \multicolumn{1}{r}{}                  & 1024                 & 1024                 \\
                                         & PubMed+PMC      & \multicolumn{1}{r}{} & \multicolumn{1}{r}{} & \multicolumn{1}{r}{}                  & 1024                 & 1024                 \\ \midrule
\multirow{7}{*}{Name-entity Recognition} & NCBI Disease    & Disease              & 6881                 & \multirow{7}{*}{Multi-Task}           & \multirow{7}{*}{512} & \multirow{7}{*}{512} \\
                                         & BC5CDR Disease  & Disease              & 19,665               &                                       &                      &                      \\
                                         & BC5CDR Chem     & Disease              & 12,694               &                                       &                      &                      \\
                                         & BC4CHEMD        & Chemical                 & 15,411               &                                       &                      &                      \\
                                         & BC2HM           & Chemical                 & 79,842               &                                       &                      &                      \\
                                         & JNLPBA          & Gene                 & 20,703               &                                       &                      &                      \\
                                         & Species-800     & Species              & 3708                 &                                       &                      &                      \\ \midrule
Relational Extraction                    & Chemprot        & Protein-chemical     & 10,031               & Single-Task                           & 256                  & 16                   \\
                                         & DDI             & Biomedical relation  & 4,920                & Single-Task                           & 256                  & 16                   \\ \midrule
Document classification                  & HoC             & Biomedical Documents & 1,580                & Single-Task                           & 256                  & 64                   \\ \midrule
Inference                                & MedNLI          & Clinical pairs       & 14,049               & Single-Task                           & 256                  & 12                   \\ \midrule
\multirow{3}{*}{Question Answering}      & BioASQ4-factoid & Biomedical QA        & 488                  & Single-Task                           & 512                  & 128                  \\
                                         & BioASQ5-factoid & Biomedical QA        & 636                  & Single-Task                           & 512                  & 128                  \\
                                         & BioASQ6-factoid & Biomedical QA        & 779                  & Single-Task                           & 512                  & 128                  \\ \bottomrule
\end{tabular}
\label{setting}
\begin{tablenotes}
      \small
      \item \textit{Notes:} The number of entities is the sum of annotations, relations, documents, pairs, and question \& answer pairs for each correspond task in the train, valid, and test sets. The statistics from \cite{DBLP:journals/corr/abs-1901-08746}, \cite{DBLP:journals/corr/abs-1906-05474}, \cite{article}, and \cite{DBLP:journals/corr/abs-1810-10566}
\end{tablenotes}
\end{table*}
\section{Results}
We tested SciFive on 7 NER tasks, 5 RE asks, 1 inference task, 1 document classification task, and 3 question answering tasks. We then compared the SciFive results with the current SOTA on these tasks.

\subsection{Data}
We describe here the datasets and the preprocessing techniques we used. In most cases, we use the same preprocessing procedure as the current baseline models (i.e. BioBERT from \cite{DBLP:journals/corr/abs-1901-08746} and BlueBERT from \cite{DBLP:journals/corr/abs-1906-05474}).

\subsubsection{Named Entity Recognition}
We tested SciFive on 7 datasets commonly used for biomedical NER: NCBI disease \citep{doan2014disease}, BC5CDR disease \citep{article_li}, BC5CDR chemical \citep{article_li}, BC4CHEMD \citep{article_krallinger}, BC2GM \citep{article_smith}, JNLPBA \citep{collier-kim-2004-introduction}, and Species800 \cite{Pafilis2013TheSA}. We follow the processing pipeline and the train/valid/test split similiar to \cite{DBLP:journals/corr/abs-1901-08746}. For all NER tasks, we evaluate the performance of SciFive based on precision (P), recall (R), and F-1 score (F).

\subsubsection{Relation Extraction}
We tested SciFive on 2 RE tasks: CHEMPROT \citep{10.1093/database/bay147} and DDI \citep{HERREROZAZO2013914}. We follow the same preprocessing technique as \cite{DBLP:journals/corr/abs-1906-05474}. We also evaluate the F1-scores of each class in the two relation extraction corpus.
\subsubsection{Natural Language Inference}
To assess the NLI capabilities of SciFive, we use the MedNLI datasets from MIMIC-III \citep{romanov-shivade-2018-lessons} with the same preprocessing technique and training/testing sets.

\subsubsection{Document Classification}
We use SciFive to classify documents from the HoC dataset \citep{10.1093/bioinformatics/btv585}, evaluating the F1 score on the sample average in the same manner as \cite{10.1093/bioinformatics/btx659}.

\subsubsection{Question Answering}
Question Answering (QA) is perhaps the most important component of our assessment, as we expect a text-to-text model to vastly outperform BERT-like models in this area. We test SciFive on the factoid questions from the BioASQ 4b, 5b, and 6b challenges \cite{article_bioasq}. To preprocess the BioASQ data, we use the same approach as \cite{DBLP:journals/corr/abs-1901-08746}. 

Using the same approach as the original T5, \citep{DBLP:journals/corr/abs-1910-10683}, SciFive converts all problems into a text-to-text format. Therefore, we cannot use the same evaluation procedure as BioBERT. \citep{DBLP:journals/corr/abs-1901-08746}. BioBERT determines the final answer for a question by taking the highest scoring answer across all the snippets of text corresponding to that question. Our model outputs a sequence of text, not a probability distribution, so we cannot determine our "best" answer in the same way as BioBERT. This key difference prevents us from evaluating strict accuracy as done by \cite{DBLP:journals/corr/abs-1901-08746}, so we evaluate only the lenient accuracy for each task. For a single question, SciFive answers questions using a sequence of text rather than probabilities for the start and end of the answer. SciFive uses each piece of context to answer that question individually. If SciFive answers correctly using one or more of the contextual snippets, we say SciFive has answered the question correctly according to the lenient accuracy metric.
 
 To evaluate our results, we rely on an expert assessment. SciFive outputs full-sentence answers that often do not correspond to the exact BioASQ answer provided for a given question, but, in many cases, these answers are still scientifically correct. For a meaningful  assessment of Q/A results, the scientific accuracy must be considered rather than the phrasing of the answer. Table \ref{qa_long} shows several examples of SciFive answers compared to BioBERT answers. It can be easily seen from these examples that SciFive provides clearer, more complete answers than BioBERT.

\subsection{Experimental Results}
In Table \ref{ner}, we show the results of SciFive compared to the SOTA approaches. For NER, RE, NLI, and documentation classification, we compare the F1 scores obtained by SciFive to the F1 scores obtained by the SOTA method pre-BioBERT, BioBERT \cite{DBLP:journals/corr/abs-1901-08746}, BlueBERT \cite{DBLP:journals/corr/abs-1906-05474}, BERT \cite{DBLP:journals/corr/abs-1810-04805}, and T5 \cite{DBLP:journals/corr/abs-1910-10683}. For the BioASQ tasks (Table \ref{qa}), we compare the lenient accuracy of base SciFive only with base T5 and base BioBERT due to the time required for thorough expert assessment. It should be noted that BioBERT was the winner of these BioASQ challenges. We achieved SOTA results on 3/7 NER tasks, 2/2 RE tasks, 1/1 NLI tasks, and 3/3 question answering tasks (Table \ref{ner}). We also achieved a near-SOTA result on the HoC document classification task. Based on these results, we emphasize the following point: SciFive (both base and large model) competitive results on classification tasks while also providing SOTA results on text generation tasks such as question-answering. This is a significant improvement over BERT-based models, which demonstrates  weaker performances on question-answering tasks.

\section{Discussion}
\begin{table}[]
\caption{Expert assessment result on Question Answering tasks (Lenient Accuracy)}
\begin{tabular}{@{}llllll@{}}
\toprule
Task      & BioBERT & T5    & \begin{tabular}[c]{@{}l@{}}SciFive\\ (PubMed+PMC)\end{tabular} & \begin{tabular}[c]{@{}l@{}}SciFive\\ (PMC)\end{tabular} & \begin{tabular}[c]{@{}l@{}}SciFive\\ (Pubmed)\end{tabular} \\ \midrule
BioAsq 4b & 57.14   & 85.06 & 87.66                                                          & 85.71                                                   & 88.31                                                      \\
BioAsq 5b & 64.83   & 86.21 & 86.21                                                          & 88.28                                                   & 88.28                                                      \\
BioAsq 6b & 57.52   & 75.82 & 75.16                                                          & 79.08                                                   & 72.55                                                      \\ \bottomrule
\end{tabular}
\label{qa}
\end{table}
We used SciFive to explore the role of text generation models in broad-spectrum biomedical NLP, achieving SOTA results on a variety of tasks. This is particularly true for question answering, where SciFive achieved SOTA results. Both T5 and SciFive significantly outperformed BioBERT, highlighting the value of text generation models in biomedical NLP. However, question answering is relatively simplistic compared to other text generation tasks. To fully examine the potential of text generation models in the context of domain-specific literature, SciFive will be applied to tasks such as document summarization and abstract generation. 

From our results, it can be seen that the SOTA results are split between the various versions of SciFive. While we expected the Pubmed+PMC model to have the best performances given the mixture of abstracts and full text articles, our results show that further study is needed to understand the optimal nature of biomedical corpora.
%
%

\section{Conclusion}

In this manuscript, we introduce SciFive, a domain-specific text-to-text model trained specifically for tasks involving biomedical literature. SciFive is effective for NER, RE, NLI, and question answering tasks, achieving SOTA or near-SOTA results in all cases. This outcome supports our conclusion that text-to-text (text generation) models are highly versatile and broadly applicable within domain-specific contexts. These models can be used for common tasks and tasks which require a longer sequence of text as an output ({\it i.e.} question answering). Our results suggest the need for further study of domain-specific text generation models applied to more difficult tasks such as a document summarization and abstract generation.

\section*{Funding}

This work has been supported by the Cancer Research Training Award (CRTA) through the National Cancer Institute to JTA and EB. This research was supported by the Intramural Research Program of the NIH.
\begin{table}[]
\caption{Example of answer generated from SciFive and BioBERT for QA tasks}
\begin{tabular}{|c|p{5cm}|p{1cm}|p{9.5cm}|}
\toprule
Task                & Question                                                                                                                        &         & Text Answer                                                                                                                                                                                                                                           \\ \midrule
\multirow{8}{*}{4b} & \multirow{2}{*}{\begin{tabular}[c]{@{}l@{}}What was the purpose of the\\ FANTOM4 project?\end{tabular}}                         & BioBert & Mammalian Genomes 4 (FANTOM4)                                                                                                                                                                                                                            \\ \cmidrule(l){4-4} 
                    &                                                                                                                                 & SciFive & \begin{tabular}[c]{@{}l@{}}the international functional annotation of the mammalian genomes 4\\ (fantom4) research collaboration set out to better understand the\\ transcriptional network that regulates macrophage differentiation\end{tabular}       \\ \cmidrule(l){2-4} 
 & \multirow{2}{*}{What is the RESID database?}                                                                                    & BioBert & RESID                                                                                                                                                                                                                                                    \\ \cmidrule(l){4-4} 
                    &                                                                                                                                 & SciFive & \begin{tabular}[c]{@{}l@{}}the resid database of protein modifications is a comprehensive collection\\ of annotations and structures for protein modifications and cross-links\\ including pre-, co-, and post-translational modifications.\end{tabular} \\ \cmidrule(l){1-4} 
\multirow{6}{*}{5b} & \multirow{2}{*}{\begin{tabular}[c]{@{}l@{}}What is the role of gamma-secreatase\\ complex in Alzheimer's Disease?\end{tabular}} & BioBert & APH-1a                                                                                                                                                                                                                                                   \\ \cmidrule(l){4-4} 
                    &                                                                                                                                 & SciFive & it cleaves a precursor to create the amyloid beta peptide                                                                                                                                                                                                \\ \cmidrule(l){2-4} 
 & \multirow{2}{*}{\begin{tabular}[c]{@{}l@{}}What is the function of BAX?\end{tabular}}            & BioBert & \begin{tabular}[c]{@{}l@{}}mitochondrial\end{tabular}                                                                                                                         \\ \cmidrule(l){4-4} 
                    &                                                                                                                                 & SciFive & \begin{tabular}[c]{@{}l@{}}bax, a central cell death regulator, is an indispensable gateway to mitochondrial \\ dysfunction and a major proapoptotic member of the b-cell lymphoma 2 (bcl-2) \\ family\end{tabular}                                                                         \\ \cmidrule(l){1-4} 
\multirow{8}{*}{6b} & \multirow{2}{*}{\begin{tabular}[c]{@{}l@{}}What is the function of the gene\\ MDA5?\end{tabular}}                               & BioBert & RIG-1                                                                                                                                                                                                                                                    \\ \cmidrule(l){4-4} 
                    &                                                                                                                                 & SciFive & \begin{tabular}[c]{@{}l@{}}melanoma differentiation-associated gene 5 (mda5) is a pattern recognition\\ receptor that recognizes cytoplasmic viral double-stranded rna (dsrna) and\\ initiates rapid innate antiviral responses.\end{tabular}            \\ \cmidrule(l){2-4} 
 & \multirow{2}{*}{\begin{tabular}[c]{@{}l@{}}What is the function of HDAC\\ proteins?\end{tabular}}                               & BioBert & Histone deacetylase                                                                                                                                                                                                                                      \\ \cmidrule(l){4-4} 
                    &                                                                                                                                 & SciFive & \begin{tabular}[c]{@{}l@{}}histone deacetylases (hdacs) prevent the relaxation of chromatin, and\\ positively or negatively regulate transcription.\end{tabular}                                                                                         \\ \bottomrule  
\end{tabular}
\label{qa_long}
\end{table}
\begin{landscape}

\begin{table}[]
\caption{Test results in biomedical named entity recognition, relation extraction, document classification, and inteference tasks}
\begin{tabular}{@{}lllllllllllllllllll@{}}
\toprule
\multicolumn{4}{l}{}                                                                                                            &         &                &                                  & \multicolumn{6}{c}{\textbf{Base}}                                                                                                                                                                                                                                              & \multicolumn{6}{c}{\textbf{Large}}                                                                                                                                                                                                                                 \\ \midrule
\multicolumn{4}{l}{}                                                                                                            & Metrics & SOTA           & \multicolumn{1}{l|}{Bert (base)} & T5    & BlueBERT   & BioBert        & \textbf{\begin{tabular}[c]{@{}l@{}}SciFive\\ (PMC\\ +PubMed)\end{tabular}} & \textbf{\begin{tabular}[c]{@{}l@{}}SciFive\\ (PubMed)\end{tabular}} & \multicolumn{1}{l|}{\textbf{\begin{tabular}[c]{@{}l@{}}SciFive\\ (PMC)\end{tabular}}} & T5          & BlueBERT      & BioBert        & \textbf{\begin{tabular}[c]{@{}l@{}}SciFive\\ (PMC\\ +PubMed)\end{tabular}} & \textbf{\begin{tabular}[c]{@{}l@{}}SciFive\\ (PubMed)\end{tabular}} & \textbf{\begin{tabular}[c]{@{}l@{}}SciFive\\ (PMC)\end{tabular}} \\ \midrule
\multirow{21}{*}{NER}                    & \multicolumn{2}{l}{\multirow{6}{*}{Disease}}      & \multirow{3}{*}{NCBI disease}    & P       &                & \multicolumn{1}{l|}{84.12}       & 87.18 & -          & {\ul 88.22}    & 88.28                                                                      & 86.28                                                               & \multicolumn{1}{l|}{\textbf{88.65}}                                                   & 87.48       & -             & 87.70          & 88.10                                                                      & 88.52                                                               & 87.64                                                            \\
                                         & \multicolumn{2}{l}{}                              &                                  & R       &                & \multicolumn{1}{l|}{87.19}       & 89.93 & -          & \textbf{91.25} & 89.30                                                                      & 89.71                                                               & \multicolumn{1}{l|}{{\ul 90.14}}                                                      & 90.14       & -             & 89.90          & 90.14                                                                      & 89.82                                                               & 89.30                                                            \\
                                         & \multicolumn{2}{l}{}                              &                                  & F       & 88.60          & \multicolumn{1}{l|}{85.63}       & 88.54 & -          & \textbf{89.71} & 88.79                                                                      & 87.96                                                               & \multicolumn{1}{l|}{{\ul 89.39}}                                                      & 88.78       & -             & 88.79          & 89.11                                                                      & 89.17                                                               & 88.46                                                            \\
                                         & \multicolumn{2}{l}{}                              & \multirow{3}{*}{BC5CDR Disease}  & P       &                & \multicolumn{1}{l|}{81.97}       & 85.95 & -          & 86.47          & 86.67                                                                      & 86.53                                                               & \multicolumn{1}{l|}{86.48}                                                            & 84.28       & -             & -              & {\ul 86.73}                                                                & 86.30                                                               & \textbf{87.01}                                                   \\
                                         & \multicolumn{2}{l}{}                              &                                  & R       &                & \multicolumn{1}{l|}{82.48}       & 87.73 & -          & 87.84          & 88.01                                                                      & \textbf{88.37}                                                      & \multicolumn{1}{l|}{87.99}                                                            & 87.38       & -             & -              & 88.46                                                                      & 87.67                                                               & {\ul 88.24}                                                      \\
                                         & \multicolumn{2}{l}{}                              &                                  & F       & 86.23          & \multicolumn{1}{l|}{82.41}       & 86.83 & 86.6       & 87.15          & 87.33                                                                      & {\ul 87.44}                                                         & \multicolumn{1}{l|}{87.23}                                                            & 86.31       & 83.8          & -              & 87.59                                                                      & 86.98                                                               & \textbf{87.62}                                                   \\ \cmidrule(l){2-19} 
                                         & \multicolumn{2}{l}{\multirow{6}{*}{Drug/chem}}    & \multirow{3}{*}{BC5CDR Chemical} & P       &                & \multicolumn{1}{l|}{90.94}       & 93.30 & -          & 93.68          & 93.89                                                                      & 94.01                                                               & \multicolumn{1}{l|}{{\ul 94.09}}                                                      & 93.44       & -             & 93.18          & \textbf{94.13}                                                             & 93.98                                                               & 93.86                                                            \\
                                         & \multicolumn{2}{l}{}                              &                                  & R       &                & \multicolumn{1}{l|}{91.38}       & 93.92 & -          & 93.26          & 94.80                                                                      & 94.69                                                               & \multicolumn{1}{l|}{94.28}                                                            & 95.02       & -             & 92.09          & \textbf{95.39}                                                             & 95.36                                                               & {\ul 95.37}                                                      \\
                                         & \multicolumn{2}{l}{}                              &                                  & F       & 93.31          & \multicolumn{1}{l|}{91.16}       & 93.61 & 93.5       & 93.47          & 94.34                                                                      & 94.35                                                               & \multicolumn{1}{l|}{94.18}                                                            & 94.22       & 91.7          & 92.63          & \textbf{94.76}                                                             & {\ul 94.66}                                                         & 94.61                                                            \\
                                         & \multicolumn{2}{l}{}                              & \multirow{3}{*}{BC4CHEMD}        & P       &                & \multicolumn{1}{l|}{91.19}       & 90.57 & -          & 92.80          & 92.50                                                                      & 92.71                                                               & \multicolumn{1}{l|}{92.01}                                                            & 91.19       & -             & \textbf{93.00} & {\ul 92.89}                                                                & 92.19                                                               & 91.98                                                            \\
                                         & \multicolumn{2}{l}{}                              &                                  & R       &                & \multicolumn{1}{l|}{88.92}       & 88.90 & -          & {\ul 91.92}    & 91.53                                                                      & 91.35                                                               & \multicolumn{1}{l|}{91.87}                                                            & 88.76       & -             & \textbf{92.35} & 91.17                                                                      & 91.73                                                               & 91.15                                                            \\
                                         & \multicolumn{2}{l}{}                              &                                  & F       & 91.14          & \multicolumn{1}{l|}{90.04}       & 89.73 & -          & {\ul 92.36}    & 92.01                                                                      & 92.02                                                               & \multicolumn{1}{l|}{92.07}                                                            & 89.96       & -             & \textbf{92.67} & 92.03                                                                      & 91.96                                                               & 91.56                                                            \\ \cmidrule(l){2-19} 
                                         & \multicolumn{2}{l}{\multirow{6}{*}{Gene/protein}} & \multirow{3}{*}{BC2GM}           & P       &                & \multicolumn{1}{l|}{81.17}       & 82.43 & -          & 84.32          & 84.44                                                                      & \textbf{84.97}                                                      & \multicolumn{1}{l|}{83.66}                                                            & 82.63       & -             & {\ul 84.78}    & 84.20                                                                      & 83.81                                                               & 83.95                                                            \\
                                         & \multicolumn{2}{l}{}                              &                                  & R       &                & \multicolumn{1}{l|}{82.42}       & 82.17 & -          & {\ul 85.12}    & 83.89                                                                      & 82.89                                                               & \multicolumn{1}{l|}{83.04}                                                            & 82.10       & -             & \textbf{85.25} & 83.48                                                                      & 83.39                                                               & 83.20                                                            \\
                                         & \multicolumn{2}{l}{}                              &                                  & F       & 81.69          & \multicolumn{1}{l|}{81.79}       & 82.29 & -          & {\ul 84.72}    & 84.16                                                                      & 83.92                                                               & \multicolumn{1}{l|}{84.29}                                                            & 82.36       & -             & \textbf{85.01} & 83.84                                                                      & 83.60                                                               & 83.57                                                            \\
                                         & \multicolumn{2}{l}{}                              & \multirow{3}{*}{JNLPBA}          & P       &                & \multicolumn{1}{l|}{69.57}       & 69.35 & -          & {\ul 72.24}    & 70.36                                                                      & 70.91                                                               & \multicolumn{1}{l|}{70.65}                                                            & 71.04       & -             & -              & 71.08                                                                      & 71.36                                                               & \textbf{77.68}                                                   \\
                                         & \multicolumn{2}{l}{}                              &                                  & R       &                & \multicolumn{1}{l|}{81.20}       & 80.61 & -          & \textbf{83.56} & 80.96                                                                      & 80.96                                                               & \multicolumn{1}{l|}{81.99}                                                            & 81.31       & -             & -              & {\ul 81.62}                                                                & 81.46                                                               & 77.42                                                            \\
                                         & \multicolumn{2}{l}{}                              &                                  & F       & \textbf{78.58} & \multicolumn{1}{l|}{74.94}       & 74.56 & -          & 77.49          & 75.29                                                                      & 75.60                                                               & \multicolumn{1}{l|}{75.89}                                                            & 75.83       & \textbf{-}    & \textbf{-}     & 75.99                                                                      & 76.08                                                               & \textbf{77.55}                                                   \\ \cmidrule(l){2-19} 
                                         & \multicolumn{2}{l}{\multirow{3}{*}{SPECIES}}      & \multirow{3}{*}{Species-800}     & P       &                & \multicolumn{1}{l|}{69.35}       & 72.18 & -          & 72.80          & 73.47                                                                      & {\ul 73.84}                                                         & \multicolumn{1}{l|}{72.68}                                                            & 72.69       & -             & -              & 72.55                                                                      & 73.08                                                               & \textbf{74.09}                                                   \\
                                         & \multicolumn{2}{l}{}                              &                                  & R       &                & \multicolumn{1}{l|}{74.05}       & 76.59 & -          & 75.36          & {\ul 79.33}                                                                & \textbf{79.45}                                                      & \multicolumn{1}{l|}{79.83}                                                            & 76.84       & -             & -              & 77.33                                                                      & 78.08                                                               & 78.71                                                            \\
                                         & \multicolumn{2}{l}{}                              &                                  & F       & 74.98          & \multicolumn{1}{l|}{71.63}       & 74.32 & -          & 74.06          & 76.29                                                                      & \textbf{76.55}                                                      & \multicolumn{1}{l|}{76.08}                                                            & 74.66       & -             & -              & 74.86                                                                      & 75.50                                                               & {\ul 76.33}                                                      \\ \midrule
\multicolumn{1}{c}{\multirow{6}{*}{RE}}  & \multicolumn{3}{l}{\multirow{3}{*}{ChemProt}}                                        & P       & 74.80          & \multicolumn{1}{l|}{76.02}       & 81    & {\ul }     & 77.02          & 82.59                                                                      & \textbf{84.24}                                                      & \multicolumn{1}{l|}{82.35}                                                            & {\ul 84.04} & -             & -              & 81.99                                                                      & 81.31                                                               & 83.58                                                            \\
\multicolumn{1}{c}{}                     & \multicolumn{3}{l}{}                                                                 & R       & 56.00          & \multicolumn{1}{l|}{71.60}       & 89.01 & {\ul }     & 75.90          & 91.21                                                                      & 93.96                                                               & \multicolumn{1}{l|}{92.31}                                                            & 86.81       & -             & -              & 95.06                                                                      & \textbf{95.60}                                                      & {\ul 95.06}                                                      \\
\multicolumn{1}{c}{}                     & \multicolumn{3}{l}{}                                                                 & F       & 64.10          & \multicolumn{1}{l|}{73.74}       & 84.82 & 72.5       & 76.46          & 86.68                                                                      & {\ul 88.83}                                                         & \multicolumn{1}{l|}{87.04}                                                            & 85.41       & 74.4          & -              & 88.04                                                                      & 87.88                                                               & \textbf{88.95}                                                   \\
\multicolumn{1}{c}{}                     & \multicolumn{3}{l}{\multirow{3}{*}{DDI}}                                             & P       & \textbf{-}     & \multicolumn{1}{l|}{\textbf{-}}  & 82.68 & \textbf{-} & \textbf{-}     & 81.96                                                                      & 83.15                                                               & \multicolumn{1}{l|}{82.75}                                                            & 83.87       & -             & \textbf{-}     & \textbf{84.22}                                                             & {\ul 83.88}                                                         & 83.00                                                            \\
\multicolumn{1}{c}{}                     & \multicolumn{3}{l}{}                                                                 & R       & \textbf{-}     & \multicolumn{1}{l|}{\textbf{-}}  & 81.41 & \textbf{-} & \textbf{-}     & 83.04                                                                      & 83.15                                                               & \multicolumn{1}{l|}{82.33}                                                            & 82.84       & -             & \textbf{-}     & 82.84                                                                      & {\ul 83.45}                                                         & \textbf{84.27}                                                   \\
\multicolumn{1}{c}{}                     & \multicolumn{3}{l}{}                                                                 & F       & 72.9           & \multicolumn{1}{l|}{\textbf{-}}  & 82.04 & 79.4       & \textbf{-}     & 82.50                                                                      & 83.15                                                               & \multicolumn{1}{l|}{82.54}                                                            & 83.35       & 79.9          & \textbf{-}     & 83.52                                                                      & \textbf{83.67}                                                      & {\ul 83.63}                                                      \\ \midrule
\multicolumn{1}{c}{\multirow{3}{*}{DoC}} & \multicolumn{3}{l}{\multirow{3}{*}{HoC}}                                             & P       & \textbf{-}     & \multicolumn{1}{l|}{\textbf{-}}  & 85.55 & \textbf{-} & \textbf{-}     & 86.27                                                                      & 86.18                                                               & \multicolumn{1}{l|}{86.08}                                                            & 86.02       & \textbf{-}    & \textbf{-}     & 86.11                                                                      & 86.35                                                               & 86.36                                                            \\
\multicolumn{1}{c}{}                     & \multicolumn{3}{l}{}                                                                 & R       & \textbf{-}     & \multicolumn{1}{l|}{\textbf{-}}  & 85.42 & \textbf{-} & \textbf{-}     & 86.29                                                                      & 86.17                                                               & \multicolumn{1}{l|}{86.20}                                                            & 85.95       & \textbf{-}    & \textbf{-}     & 86.21                                                                      & 86.31                                                               & 86.39                                                            \\
\multicolumn{1}{c}{}                     & \multicolumn{3}{l}{}                                                                 & F*      & 81.5           & \multicolumn{1}{l|}{\textbf{-}}  & 85.22 & 85.3       & \textbf{-}     & 85.99                                                                      & 85.89                                                               & \multicolumn{1}{l|}{85.83}                                                            & 85.68       & \textbf{87.3} & \textbf{-}     & 85.87                                                                      & 86.03                                                               & {\ul 86.08}                                                      \\ \midrule
\multicolumn{1}{c}{NLI}                  & \multicolumn{3}{l}{MedNLI}                                                           & Acc     & 73.5           & \multicolumn{1}{l|}{\textbf{-}}  & 83.90 & 84.0       & \textbf{-}     & 84.88                                                                      & 85.30                                                               & \multicolumn{1}{l|}{84.25}                                                            & 83.8        & 83.8          & \textbf{-}     & \textbf{86.57}                                                             & {\ul 86.36}                                                         & 86.08                                                            \\ \bottomrule
\end{tabular}
\label{ner}

\begin{tablenotes}
      \small
      \item \textit{Notes:} P for Precision; R for Recall; F for F1 score; F* is F1 score on sample average. Best scores are in bold, second best scores are underlined. Baseline result and SOTA from \cite{DBLP:journals/corr/abs-1901-08746} and \cite{DBLP:journals/corr/abs-1906-05474}
\end{tablenotes}

\end{table}
\end{landscape}

%
%

\bibliographystyle{natbib}
\bibliography{references}

\begin{thebibliography}{}

\bibitem[Baker {\em et~al.}(2015)Baker, Silins, Guo, Ali, Högberg, Stenius,
  and Korhonen]{10.1093/bioinformatics/btv585}
Baker, S., Silins, I., Guo, Y., Ali, I., Högberg, J., Stenius, U., and
  Korhonen, A. (2015).
\newblock {Automatic semantic classification of scientific literature according
  to the hallmarks of cancer}.
\newblock {\em Bioinformatics\/}, {\bf 32}(3), 432--440.

\bibitem[Collier and Kim(2004)Collier and Kim]{collier-kim-2004-introduction}
Collier, N. and Kim, J.-D. (2004).
\newblock Introduction to the bio-entity recognition task at {JNLPBA}.
\newblock In {\em Proceedings of the International Joint Workshop on Natural
  Language Processing in Biomedicine and its Applications
  ({NLPBA}/{B}io{NLP})\/}, pages 73--78, Geneva, Switzerland. COLING.

\bibitem[Devlin {\em et~al.}(2018)Devlin, Chang, Lee, and
  Toutanova]{DBLP:journals/corr/abs-1810-04805}
Devlin, J., Chang, M., Lee, K., and Toutanova, K. (2018).
\newblock {BERT:} pre-training of deep bidirectional transformers for language
  understanding.
\newblock {\em CoRR\/}, {\bf abs/1810.04805}.

\bibitem[Dodge {\em et~al.}(2021)Dodge, Sap, Marasovic, Agnew, Ilharco,
  Groeneveld, and Gardner]{DBLP:journals/corr/abs-2104-08758}
Dodge, J., Sap, M., Marasovic, A., Agnew, W., Ilharco, G., Groeneveld, D., and
  Gardner, M. (2021).
\newblock Documenting the english colossal clean crawled corpus.
\newblock {\em CoRR\/}, {\bf abs/2104.08758}.

\bibitem[Doğan {\em et~al.}(2014)Doğan, Leaman, and Lu]{doan2014disease}
Doğan, R.~I., Leaman, R., and Lu, Z. (2014).
\newblock Ncbi disease corpus: A resource for disease name recognition and
  concept normalization.
\newblock {\em Journal of Biomedical Informatics\/}, {\bf 47}, 1 -- 10.

\bibitem[Habibi {\em et~al.}(2017)Habibi, Weber, Neves, Wiegandt, and
  Leser]{article}
Habibi, M., Weber, L., Neves, M., Wiegandt, D., and Leser, U. (2017).
\newblock Deep learning with word embeddings improves biomedical named entity
  recognition.
\newblock {\em Bioinformatics (Oxford, England)\/}, {\bf 33}, i37--i48.

\bibitem[Herrero-Zazo {\em et~al.}(2013)Herrero-Zazo, Segura-Bedmar, Martínez,
  and Declerck]{HERREROZAZO2013914}
Herrero-Zazo, M., Segura-Bedmar, I., Martínez, P., and Declerck, T. (2013).
\newblock The ddi corpus: An annotated corpus with pharmacological substances
  and drug–drug interactions.
\newblock {\em Journal of Biomedical Informatics\/}, {\bf 46}(5), 914--920.

\bibitem[Islamaj~Doğan {\em et~al.}(2019)Islamaj~Doğan, Kim,
  Chatr-aryamontri, Wei, Comeau, Antunes, Matos, Chen, Elangovan, Panyam,
  Verspoor, Liu, Wang, Liu, Altınel, Hüsünbeyi, Özgür, Fergadis, Wang,
  Dai, Tran, Kavuluru, Luo, Steppi, Zhang, Qu, and Lu]{10.1093/database/bay147}
Islamaj~Doğan, R., Kim, S., Chatr-aryamontri, A., Wei, C.-H., Comeau, D.~C.,
  Antunes, R., Matos, S., Chen, Q., Elangovan, A., Panyam, N.~C., Verspoor, K.,
  Liu, H., Wang, Y., Liu, Z., Altınel, B., Hüsünbeyi, Z.~M., Özgür, A.,
  Fergadis, A., Wang, C.-K., Dai, H.-J., Tran, T., Kavuluru, R., Luo, L.,
  Steppi, A., Zhang, J., Qu, J., and Lu, Z. (2019).
\newblock {Overview of the BioCreative VI Precision Medicine Track: mining
  protein interactions and mutations for precision medicine}.
\newblock {\em Database\/}, {\bf 2019}.
\newblock bay147.

\bibitem[Krallinger {\em et~al.}(2015)Krallinger, Rabal, Leitner, Vazquez,
  Salgado, lu, Leaman, Lu, Ji, Lowe, Sayle, Batista-Navarro, Rak, Huber,
  Rocktäschel, Matos, Campos, Tang, Xu, and Valencia]{article_krallinger}
Krallinger, M., Rabal, O., Leitner, F., Vazquez, M., Salgado, D., lu, Z.,
  Leaman, R., Lu, Y., Ji, D., Lowe, D., Sayle, R., Batista-Navarro, R., Rak,
  R., Huber, T., Rocktäschel, T., Matos, S., Campos, D., Tang, B., Xu, H., and
  Valencia, A. (2015).
\newblock The chemdner corpus of chemicals and drugs and its annotation
  principles.
\newblock {\em Journal of Cheminformatics\/}, {\bf 7}, S2.

\bibitem[Kudo and Richardson(2018)Kudo and
  Richardson]{DBLP:journals/corr/abs-1808-06226}
Kudo, T. and Richardson, J. (2018).
\newblock Sentencepiece: {A} simple and language independent subword tokenizer
  and detokenizer for neural text processing.
\newblock {\em CoRR\/}, {\bf abs/1808.06226}.

\bibitem[Lee {\em et~al.}(2019)Lee, Yoon, Kim, Kim, Kim, So, and
  Kang]{DBLP:journals/corr/abs-1901-08746}
Lee, J., Yoon, W., Kim, S., Kim, D., Kim, S., So, C.~H., and Kang, J. (2019).
\newblock Biobert: a pre-trained biomedical language representation model for
  biomedical text mining.
\newblock {\em CoRR\/}, {\bf abs/1901.08746}.

\bibitem[Li {\em et~al.}(2016)Li, Sun, Johnson, Sciaky, Wei, Leaman, Davis,
  Mattingly, Wiegers, and lu]{article_li}
Li, J., Sun, Y., Johnson, R., Sciaky, D., Wei, C.-H., Leaman, R., Davis, A.~P.,
  Mattingly, C., Wiegers, T., and lu, Z. (2016).
\newblock Biocreative v cdr task corpus: a resource for chemical disease
  relation extraction.
\newblock {\em Database\/}, {\bf 2016}, baw068.

\bibitem[Pafilis {\em et~al.}(2013)Pafilis, Frankild, Fanini, Faulwetter,
  Pavloudi, Vasileiadou, Arvanitidis, and Jensen]{Pafilis2013TheSA}
Pafilis, E., Frankild, S., Fanini, L., Faulwetter, S., Pavloudi, C.,
  Vasileiadou, A., Arvanitidis, C., and Jensen, L. (2013).
\newblock The species and organisms resources for fast and accurate
  identification of taxonomic names in text.
\newblock {\em PLoS ONE\/}, {\bf 8}.

\bibitem[Peng {\em et~al.}(2019)Peng, Yan, and
  Lu]{DBLP:journals/corr/abs-1906-05474}
Peng, Y., Yan, S., and Lu, Z. (2019).
\newblock Transfer learning in biomedical natural language processing: An
  evaluation of {BERT} and elmo on ten benchmarking datasets.
\newblock {\em CoRR\/}, {\bf abs/1906.05474}.

\bibitem[Raffel {\em et~al.}(2019)Raffel, Shazeer, Roberts, Lee, Narang,
  Matena, Zhou, Li, and Liu]{DBLP:journals/corr/abs-1910-10683}
Raffel, C., Shazeer, N., Roberts, A., Lee, K., Narang, S., Matena, M., Zhou,
  Y., Li, W., and Liu, P.~J. (2019).
\newblock Exploring the limits of transfer learning with a unified text-to-text
  transformer.
\newblock {\em CoRR\/}, {\bf abs/1910.10683}.

\bibitem[Romanov and Shivade(2018)Romanov and
  Shivade]{romanov-shivade-2018-lessons}
Romanov, A. and Shivade, C. (2018).
\newblock Lessons from natural language inference in the clinical domain.
\newblock In {\em Proceedings of the 2018 Conference on Empirical Methods in
  Natural Language Processing\/}, pages 1586--1596, Brussels, Belgium.
  Association for Computational Linguistics.

\bibitem[Ruder(2017)Ruder]{DBLP:journals/corr/Ruder17a}
Ruder, S. (2017).
\newblock An overview of multi-task learning in deep neural networks.
\newblock {\em CoRR\/}, {\bf abs/1706.05098}.

\bibitem[Smith {\em et~al.}(2008)Smith, Tanabe, Ando, Kuo, Chung, Hsu, Lin,
  Klinger, Friedrich, Ganchev, Torii, Liu, Haddow, Struble, Povinelli, Vlachos,
  Baumgartner~Jr, Hunter, Carpenter, and Wilbur]{article_smith}
Smith, L., Tanabe, L., Ando, R., Kuo, C., Chung, I.-F., Hsu, C., Lin, Y.,
  Klinger, R., Friedrich, C., Ganchev, K., Torii, M., Liu, H., Haddow, B.,
  Struble, C., Povinelli, R., Vlachos, A., Baumgartner~Jr, W., Hunter, L.,
  Carpenter, B., and Wilbur, W. (2008).
\newblock Overview of biocreative ii gene mention recognition.
\newblock {\em Genome Biology\/}, {\bf 9}.

\bibitem[Tsatsaronis {\em et~al.}(2015)Tsatsaronis, Balikas, Malakasiotis,
  Partalas, Zschunke, Alvers, Weißenborn, Krithara, Petridis,
  Polychronopoulos, Almirantis, Pavlopoulos, Baskiotis, Gallinari, Artieres,
  Ngonga~Ngomo, Heino, Gaussier, Barrio-Alvers, and Paliouras]{article_bioasq}
Tsatsaronis, G., Balikas, G., Malakasiotis, P., Partalas, I., Zschunke, M.,
  Alvers, M., Weißenborn, D., Krithara, A., Petridis, S., Polychronopoulos,
  D., Almirantis, Y., Pavlopoulos, J., Baskiotis, N., Gallinari, P., Artieres,
  T., Ngonga~Ngomo, A.-C., Heino, N., Gaussier, E., Barrio-Alvers, L., and
  Paliouras, G. (2015).
\newblock An overview of the bioasq large-scale biomedical semantic indexing
  and question answering competition.
\newblock {\em BMC Bioinformatics\/}, {\bf 16}, 138.

\bibitem[Vaswani {\em et~al.}(2017)Vaswani, Shazeer, Parmar, Uszkoreit, Jones,
  Gomez, Kaiser, and Polosukhin]{DBLP:journals/corr/VaswaniSPUJGKP17}
Vaswani, A., Shazeer, N., Parmar, N., Uszkoreit, J., Jones, L., Gomez, A.~N.,
  Kaiser, L., and Polosukhin, I. (2017).
\newblock Attention is all you need.
\newblock {\em CoRR\/}, {\bf abs/1706.03762}.

\bibitem[Zhang {\em et~al.}(2017)Zhang, Zheng, Lin, Wang, Yang, and
  Dumontier]{10.1093/bioinformatics/btx659}
Zhang, Y., Zheng, W., Lin, H., Wang, J., Yang, Z., and Dumontier, M. (2017).
\newblock {Drug–drug interaction extraction via hierarchical RNNs on sequence
  and shortest dependency paths}.
\newblock {\em Bioinformatics\/}, {\bf 34}(5), 828--835.

\bibitem[Zhu {\em et~al.}(2018)Zhu, Paschalidis, and
  Tahmasebi]{DBLP:journals/corr/abs-1810-10566}
Zhu, H., Paschalidis, I.~C., and Tahmasebi, A. (2018).
\newblock Clinical concept extraction with contextual word embedding.
\newblock {\em CoRR\/}, {\bf abs/1810.10566}.

\end{thebibliography}








\end{document}